\documentclass{esannV2}
\usepackage{graphicx}
\usepackage[utf8]{inputenc}
\usepackage{amssymb,amsmath,array}
\usepackage[english]{babel}

\begin{document}
\title{Search Strategies for Binary Feature Selection for a Naive Bayes Classifier}

\author{Tsirizo Rabenoro$^{1,2}$, Jérôme Lacaille$^1$, Marie Cottrell$^2$, and
  Fabrice Rossi$^2$
\thanks{T. Rabenoro is supported by a grant from Snecma, Safran Group.}
\vspace{.3cm}\\
1- SAMM EA 4543, Université Paris 1 Panthéon-Sorbonne\\
90, rue de Tolbiac, 75634 Paris cedex 13, France
\vspace{.1cm}\\
2- Snecma, Groupe Safran,77550 Moissy Cramayel, France\\
}

\maketitle

\begin{abstract}
  We compare in this paper several feature selection methods for the Naive
  Bayes Classifier (NBC) when the data under study are described by a large
  number of redundant binary indicators. Wrapper approaches guided by the NBC
  estimation of the classification error probability outperform filter
  approaches while retaining a reasonable computational cost. 
\end{abstract}

\section{Introduction}\label{sec:introduction}
We consider in this paper application contexts in which a large body of expert
knowledge is available as a series of simple and low level parametric
scores. The goal is to build an interpretable classifier from this knowledge
(and from a learning set). Then the scores are assumed to be simple parametric
functions from the data space to $\mathbb{R}$ with the interpretation that a
high value of a score indicates that the datum submitted to the score belongs
probably to the class that the score has been designed to detect. 

Let's consider a concrete example from our main application domain, aircraft
engine monitoring (see \cite{rabenorolacailleetal2014anomaly-detection-wcci}
for details). We aim here at classifying some short time series (around 150
time points, each series having its own specific length) into different
classes (normal signal and different types of anomalies corresponding to some
non stationarity in the signal). Domain experts have selected a set of
statistical tests as scores. For instance, the Mann-Whitney $U$ test can be
used to reject the null hypothesis that two populations are identical. It can
be applied to a time series by selecting a potential break point in the series
$t_b$ and a window size $w$, and by considering the $w/2$ points \emph{before}
$t_b$ as the first population and the $w/2$ points \emph{after} $t_b$ as the
second population. The score is the $p$-value of the U test applied to those
populations: a high value leads to not rejecting the null hypothesis and thus
is an indication that the time series belongs to ``no anomaly'' class. Notice
that the parameters of the score are here the potential break point $t_b$ and
the window size $w$. See \cite{rabenorolacailleetal2014anomaly-detection-wcci}
for other examples.

While the experts can design scores, they are seldom able to provide more than
hints about the parameters and the thresholds (i.e. when to consider that the
score is ``high enough''). In addition, the scores are generally no sufficient
alone and several of them should be combined to achieve acceptable
classification rates. We proposed in
\cite{rabenorolacailleetal2014anomaly-detection-wcci} to address this problem
via feature selection using the filter mRMR approach
\cite{peng2005feature}. The main idea, recalled in Section
\ref{sec:from-scores-binary}, consists in turning the scores into a large set
of redundant binary features and in using a feature selection method to keep
useful ones. This has the effect of finding good parameters and thresholds for
the scores while retaining a reasonable number of scores, easing
interpretation of the decision made by a Naive Bayes Classifier (NBC) built on
them. In \cite{rabenorolacailleetal2014anomaly-detection-wcci}, the selection
is done by a filter method. We investigate in the present paper wrapper based
approaches. 

\section{From scores to binary indicators}\label{sec:from-scores-binary}
We assume given a training set $(X_i,Y_i)_{1\leq i\leq N}$ where the
observations space $\mathcal{X}$ can be arbitrary while the target space is a
finite set of classes, $\mathcal{Y}=\{1,\ldots, K\}$. We are also given a set of
$Q$ parametric scores, $(s_q)_{1\leq q\leq Q}$. Each $s_q$ is a function from
$\mathcal{X}\times W_q$ to $\mathbb{R}$, where $W_q$ is the parameter space of
the score. 

The main constraint of our context is that experts only allow score results to
be used by the classifier. In addition, the semantic of the scores means that
only decisions of the form $s_q(X_i,w_q)\leq \lambda_q$ are really
meaningful. We propose to transform this set of scores into a much
larger set of binary indicators. This is done by choosing for each score a
finite subset of $W_q$, $\{w^1_{q},\ldots,w^{p_q}_{q}\}$ and a finite set of
thresholds $\{\lambda_q^1,\ldots,\lambda_q^{t_q}\}$, and by defining
$p_q\times t_q$ indicator functions by
$\mathbb{I}^{p,t}_{q}(X)=\mathbb{I}_{s_q(X,w^p_q)\leq \lambda_q^t}$. This can
be seen as a form of grid search in the ``score space'' in the sense that
tuning the scores to the data set can be done indirectly by selecting relevant
binary indicators (a similar principle is used in
e.g. \cite{fleuret-2004}). By feeding the training set $(X_i,Y_i)_{1\leq i\leq
  N}$ through the indicators, we obtain a new training set $(B_i,Y_i)_{1\leq
  i\leq N}$ where the $B_i$ take values into $\{0,1\}^{P}$ where $P$ is the
total number of binary indicators generated from the scores.

Contrarily to arbitrary binary valued variables, those indicators are
intrinsically highly redundant and correlated. For instance, if
$\lambda_q^t<\lambda_q^{t'}$, then $\mathbb{I}^{p,t}_{q}(X)=1$ implies
$\mathbb{I}^{p,t'}_{q}(X)=1$. This has adverse consequences on feature
selection methods and on the Naive Bayes Classifier. 

\section{Naive Bayes Classifier}
The Naive Bayes Classifier (NBC, this e.g. \cite{MurphyML2012}) is a very
simple and robust classifier based on the (unrealistic) assumption that
the features used to describe the objects to classify are 
conditionally independent given the class. In our context, this translates
into $P(B=b|Y=k)=\prod_{p=1}^PP(B^p=b^p|Y=k),$ where $B^p$ is the $p$-th indicator
value in the indicator vector. This allows to estimate easily the posterior
probability $P(Y=k|B=b)$ as
\[
P(Y=k|B=b)=\frac{P(Y=k)\prod_{p=1}^PP(B^p=b^p|Y=k)}{P(B^p=b)},
\]
using estimated values of the $P(B^p=b^p|Y=k)$. Those values are obtained by
simple class conditional counts. 

In our context, the motivation for using the NBC is twofold. Its classical
properties (simplicity, robustness and good performances) are of course a
first motivation (it was used successfully in e.g. \cite{fleuret-2004} with
binary indicators). In addition, the actual classification is performed in a
way that is very easy to interpret by a domain expert with limited machine
learning expertise. It consists indeed in comparing posterior probabilities,
which can be done on an indicator by indicator basis, by computing
$\frac{P(B^p=b^p|Y=k)}{P(B^p=b^p|Y=k')}$.

This allows to show to the user the indicators, and thus the underlying
scores, that are the most important in one decision, by being the more
discriminant between classes for a given observation (see
\cite{poulin2006visual} for a complete visual solution). In our application
context, a black box decision model is unacceptable, while this kind of grey
box decision is accepted by the domain experts, as long as the NBC is
constructed from their scores. 

\section{Feature selection for the NBC}
Selecting good features is of utmost importance to get good performances with
a NBC. In addition, the type of decision analysis we mentioned in the previous
section is only realistic if the number of features remains relatively small. 

Numerous feature selection methods have been investigated for NBC, ranging
from wrapper forward search \cite{DBLP:conf/uai/LangleyS94} to mRMR like
filter approach as in
\cite{fleuret-2004,rabenorolacailleetal2014anomaly-detection-wcci}, but more
sophisticated search strategies (such as forward backward methods, see
\cite{guyon2006}, chapter 4) have not. In addition, our specific context of
highly redundant binary indicators remains also unexplored. It finally be noted
that the solution recommended in text books remains a basic mutual information
based filter approach (see e.g. \cite{MurphyML2012}).

\subsection{Incremental calculation}
The main motivation of filter approaches is generally the large computational
cost of wrapper solutions, as the latter tend to give better feature subsets
than the former. Fortunately, the NBC structure allows one to implement
forward or backward strategies in a rather efficient way. Indeed, the decision
of a NBC is done by comparing posterior probabilities, which can be done
equivalently by comparing the log likelihoods of the pair $(B,k)$, for the
different classes $k$:
\[
\log P(B=b,Y=k)=\log P(Y=k)+\sum_{p=1}^P\log P(B^p=b^p|Y=k).
\]
Given a feature subset of size $m\leq P$, this can be computed in $O(Nm)$ for all
the $(B_i,Y_i)$ provided the full conditional distribution $P(B^p|Y=k)$ have
been already computed (this is done in $O(NP)$ if $K$ is small compared to
$P$). Then the effect of adding or removing a feature can be computed by
simply adding or subtracting to $\log P(B=b,Y=k)$ the contribution of the
feature, that is in a total time in $O(N)$. Then evaluating all the features
in a forward or backward step is in $O(NP)$ and thus the total cost of a
forward (or backward) search is in $O(NP^2)$. Notice that this does not apply
to arbitrary search strategies where one can move from a feature subset to a
completely different one (such as in genetic algorithms). 

This is still an order of magnitude more expensive than e.g. a simple Mutual
Information (MI) based forward search which costs $O(NP)$ but it is comparable
to mRMR which costs also $O(NP^2)$ when used to rank all the
features. Compared to classifiers for which no incremental solution exists,
forward or backward wrapper based approaches are then more affordable for the
NBC. Indeed non incremental classifiers have generally at least a training
cost in $O(Nm)$ for $m$ features, leading to a $O(NP^3)$ total cost.

\subsection{Other NBC specific aspects}
In addition to the reasonable complexity of its wrapper solutions, the NBC has
some specific aspects that should be taken into account during feature
selections. Firstly, the NBC is non monotonic as adding features can degrade
(at lot) its performances (mostly because it cannot weight features). Thus
branch-and-bound solutions cannot be used reliably.

A second issue is the evaluation metric to be used during the search. In
wrapper approaches, one uses in general the risk under consideration, that is
the classification error in our case. However, because of the additive
nature of the NBC, many features have either no effect or an identical one
when they are added one by one to an existing set of features. In other words,
the classification error is not sensitive enough to distinguish between
some of the features. We propose therefore to use the error probability as
estimated by the NBC itself as the quality measure during feature
selection. For a feature subset $S$, this is
$\frac{1}{N}\sum_{i=1}^N\hat{P}_S(Y\neq Y_i|B=b)$
where $\hat{P}_S(Y\neq Y_i|B=b)$ is the conditional probability estimated
with the features from $S$ by the NBC. Using this value amounts to taking into
account the uncertainty in the decision as estimated by the NBC itself. It
should be noted that the log conditional likelihood $\sum_{i=1}^N\log
\hat{P}_S(Y=Y_i|B=b)$ gives very poor results in this NBC context because of
the conditional independence assumption. (results are not reported here
for space reasons.)

\section{Experimental evaluation}
We compare in this section several feature selection scheme for the NBC used
on the binary indicators obtained as explained in Section
\ref{sec:from-scores-binary}. 

\subsection{Data sets}
We use simulated data sets similar to the one used in
\cite{rabenorolacailleetal2014anomaly-detection-wcci}. The training set and
the test set have identical characteristics: they are made of 6000 times
series, with 3000 normal examples (Gaussian white noise with $\sigma=1$
standard deviation) and 3000 abnormal examples belong to three different
classes (in equal proportion). The mean change anomaly consists in switching
from a $\mu=0$ mean white noise to a $\mu\in[1,5]$ white noise. The variance
change anomaly consists in switching from a $\sigma=1$ standard deviation
white noise to a $\sigma\in[2,6]$ white noise. Finally, the trend shift
anomaly adds a linear trend to the signal from the change point to the end of
the time series, with a final trend amplitude in $[1,5]$. Signal lengths are
chosen uniformly at random in $[100,200]$ time steps, while the change point
happens in the 60 \% central area of the signal (e.g. in $[20,80]$ for a
length 100 signal). 

The scores are based on sliding windows on which two population tests are
conducted (as explained in Section \ref{sec:introduction}). We use the U test,
the Kolmogorov-Smirnov test and the F-test (variance test). The parameter is
in all cases the window length. We use also confirmatory scores based on
successive windows. Details on the parameter values and on the confirmation
scores can be found in
\cite{rabenorolacailleetal2014anomaly-detection-wcci}. After the binarization
process described in Section \ref{sec:from-scores-binary} has been applied, we
obtain in this context 814 indicators. 

\subsection{General procedure and evaluation}
For each feature selection method, the NBC is built on half of the training set
(keeping class proportions) and the best feature subset is selected
using the second half of the training set (by choosing the smallest subset
among those that have the lowest classification error). The feature subset
is then evaluated on the test set by reporting the classification error. 

\subsection{Feature selection techniques}
We use two filter procedures as reference, namely a simple Mutual Information
(MI) feature ranking, and the mRMR ranking \cite{peng2005feature}. All the
other methods are wrapper approaches using either the classification error
or the error probability as performance measure. We compare a forward search
(at each step, the best feature is added to the feature set), a backward
search (at each step, the worst feature is removed from the feature set) and
full forward/backward search (also called floating search in
\cite{guyon2006}). In those algorithms, a forward phase is followed by a
backward phase (and vice versa) until the results do not improve. For
instance, one starts by a backward search to find a first optimal subset, then
proceeds to a forward search from this subset to get a better one (with more
variables). In case of improvement, the procedure is restarted from the last
subset (backward, then forward, etc.). 

\section{Results and discussion}
Results are summarized in table \ref{tab:results}. As expected, the wrapper
approaches outperform the filter ones. In addition, the high redundancy of the
binary indicators, implied by their constructions, has strong adverse effects
on the MI filter method as it tends to select very redundant indicators. The
mRMR ranking avoids this effect but obtains sub optimal results. 

\begin{table}[htb]\small
  \begin{center}
  \begin{tabular}{llcc}
Method & Perf. Measure & \# of features & test error \\\hline
MI filter & Error & 422 & 0.1387\\
mRMR filter & Error &19 & 0.1435\\\hline
Forward search & Error & 136&0.1237\\
Forward search & Probability & 207&0.1225\\
Backward search & Error & 27 & 0.1308\\
Backward search & Probability & 86 & 0.1283\\\hline
Forward--Backward & Error & 92&0.1238\\
Forward--Backward & Probability & 123&0.1237\\
Backward--Forward & Error & 112 & 0.1267\\
Backward--Forward & Probability & 122 & 0.1168\\\hline
  \end{tabular}
\end{center}
\vspace{-1.3em}
\caption{Classification error obtained on the test set }
\label{tab:results}
\end{table}

Using the error probability rather than the classification error always
improves the results of the wrapper approaches. It allows a more accurate
ordering of the features than cannot be inferred by the search procedure
alone. 

Moving from a simple greedy search to a floating search improves the
performances in the case of the backward search as it tends to select too few
variables. In the case of the forward search, the performances are slightly
degraded but the number of features is strongly reduced. 

Those results show that the filter approaches should be avoided for the
NBC. They also show that the error probability should be used to guide the
greedy search in order to get a more accurate ordering of the features at each
step of the search. The greedy search wrapper procedures give rather
comparable results with slightly increased performances for the floating
search. In the highly redundant binary indicators context, the backward
floating search guided by the error probability appears as the best solution,
contrarily to the classical recommendations for the NBC (namely, using a MI
filter). 

\begin{footnotesize}
\bibliographystyle{abbrv}
\bibliography{binary-indicators}
 \end{footnotesize}

\end{document}